\documentclass[VANCOUVER,STIX1COL]{WileyNJD-v2}
\usepackage{moreverb}

\newcommand\BibTeX{{\rmfamily B\kern-.05em \textsc{i\kern-.025em b}\kern-.08em
T\kern-.1667em\lower.7ex\hbox{E}\kern-.125emX}}

\articletype{Original Research Paper}%

\received{<day> <Month>, <year>}
\revised{<day> <Month>, <year>}
\accepted{<day> <Month>, <year>}

\begin{document}

\title{Efficient Learning of Robust Quadruped Bounding using Pretrained Neural Networks}

\author[1]{Zhicheng Wang**}

\author[1]{Anqiao Li**}

\author[1]{Yixiao Zheng}

\author[2]{Anhuan Xie}

\author[3]{Zhibin Li}

\author[1,4]{Jun Wu}

\author[1,4]{Qiuguo Zhu*}

\authormark{AUTHOR ONE \textsc{et al}}

\address[1]{\orgdiv{** The authors contributed equally to this work}}

\address[2]{\orgdiv{Institute of Cyber-Systems and Control}, \orgname{Zhejiang University}, \orgaddress{\state{Zhejiang}, \country{China}}}

\address[3]{\orgdiv{Intelligent Robot Research Center}, \orgname{Zhejiang Lab}, \orgaddress{\state{Zhejiang}, \country{China}}}

\address[4]{\orgdiv{Department of Computer Science}, \orgname{University College London}, \country{United Kingdom}}

\address[5]{\orgdiv{State Key Laboratory of Industrial Control Technology}, \orgname{Zhejiang University}, \orgaddress{\state{Zhejiang}, \country{China}}}

\corres{*Qiuguo Zhu, \email{qgzhu@zju.edu.cn}}

\presentaddress{No.38 Zheda Rd, Hangzhou, China.}

\abstract[Abstract]{
Bounding is one of the important gaits in quadrupedal locomotion for negotiating obstacles. We proposed an effective approach that can learn robust bounding gaits more efficiently despite its large variation of dynamic body movements. We first pretrained the neural network (NN) based on data from a robot which operated by conventional model-based controllers, and then we optimized further the pretrained NN via deep reinforcement learning (DRL). In particular, we designed a reward function considering contact points and phases to enforce the gait symmetry and periodicity, which improved the bounding performance. The NN-based feedback controller was learned in simulation and directly deployed on the real quadruped robot Jueying Mini successfully. We present a variety of  environments in both indoor and outdoor with our approach. Our approach shows efficient computing and good locomotion results by the Jueying Mini quadrupedal robot bounding over uneven terrain.   }


\jnlcitation{\cname{%
\author{Zhicheng W.}, 
\author{A. Li}, 
\author{Y. Zheng}, 
\author{A. Xie}, 
\author{Z. Li}, 
\author{J. Wu}, and 
\author{Q. Zhu}} (\cyear{2022}), 
\ctitle{Efficient Learning of Robust Quadruped Bounding using Pretrained Neural Networks}, \cjournal{IET Cyber-Systems and Robotics}, \cvol{2022;00:1--6}.}

\maketitle

\section{Introduction}\label{sec1}

Legged robots have been attracting more attention in recent years for their versatile motion capabilities. The motion planning and control of a legged robot has been well researched. Methods based on reduce-order models are proved to be feasible to generate adaptive gaits for real robots using prior knowledge and fine-tuning. In search of higher generalizing performance and agility , learning-based approaches, such as the deep reinforcement learning (DRL), has gained new trends in legged locomotion control to solve these problems, because it allows learning a MIMO (multiple input and multiple output) feedback control policies that can run in real-time, particularly dealing with very high dimensional sensory inputs. 

Constrained by the capability of computing devices and legged robots, DRL has not been applied to motion control for quadruped robots until recent years. Iscen\cite{iscen2018TG} achieved multiple gaits including running and bounding with the help of a predefined trajectory generator. Haarnoja and Tan \cite{haarnoja2018learning, tan2018simtoreal} first implemented DRL-trained walking, trotting, and galloping on a real Minitaur robot and verified the feasibility of the end-to-end route. Subsequently, Ha \cite{ha2020learning} utilized an on-robot DRL method with minimal human interference by constructing a physical reset mechanism quite similar to that of a computer simulation, and achieved trotting and walking on unstructured terrain. The work in \cite{hutter2016anymal, Hwangbo_2019, lee2019robust} presents learning separate skills such as trotting and fall recovery using an end-to-end DRL framework. 

 When facing discrete terrain, Tsounis \cite{tsounis2020deepgait} introduced DeepGait, a gait planner with a metachronal gait. With the help of a Timed Convolutional Network and privilege imitation, the robot conquered a series of challenging terrains using trotting \cite{Lee_2020}. To imitate real quadruped animals to the greatest extent, Peng \cite{2016-TOG-deepRL, Peng_2017, peng2020learning} collected locomotion data from real dogs and achieved trotting and spinning in robots by domain adaption. To go beyond learning of single-skill policies, multiple expert locomotion skills can be synthesized and generated within one framework, which can fuse multiple neural networks into a newly synthesized expert network \cite{yang2020multi, yuanmulti2022}.

Since based on DRL, researchers have to face the reward hacking problem, an insecure situation that agent obtains reward in an unexpected way. One of its reasons is that the optimization randomly falls into a local optimum other than the expected one\cite{ConcreteProblems}. The most common way to cope is to add specialized reward and tune manually\cite{hutter2016anymal}, which partially neutralizes the advantage of learning-based methods. Introducing predefined reference motion and imitation\cite{peng2020learning} can be another solution, which seeks a subtle balance between agility and constrains.

Most papers take trotting as demonstration task. In addition to trot and gallop, bounding is also an important form of legged locomotion. It can be used to cross different obstacles and also as a transition model between trotting and galloping \cite{Haynes-2006-9473}. However, bounding is harder to train by end-to-end DRL than trotting and galloping for several reasons. However, in bounding gait, the center of mass (CoM) and pitch angle of the robot change more violently, which usually causes termination. Galloping and trotting outweigh bounding in stability respectively under high and low speed condition. As a form of reward hacking, bounding can be easily overrode to prevent falling\cite{iscen2018TG}. Hence, quadruped bounding is rarely learned and shown in details in related works, but serve as a proof of other features such as behavior generalizing\cite{iscen2018TG}.



Our work studied an effective solution to train an end-to-end reference-free neural network controller by DRL that can perform symmetric bounding gait and can be successfully transferred to the real robot. The main contributions are as follows: 
\begin{enumerate}
\item We proposed a pre-fitting method to initialize the weights of a warm-start policy trained with data collected from model-based policy, which prevents reward hacking and behavior overriding.
\item We proposed an effective reward function based on contact phases which well regulate a periodic gait and resolves large variance and subsequent divergence in training because of large fluctuations in the body movements during bounding gaits.
\end{enumerate}
 
The organization of this paper is as follows: In Section II, we introduce the overall structure of our pre-fitting method and DRL workflow. In Section III, we introduced the platform on that we deployed the alorithms. In Section IV, we validate the feasibility of policy trained with our method on a physical robot and compare the results with those of conventional controllers. Finally, in Section V, the conclusion, along with inspiration for future work, is stated.

\section{Method}

\begin{figure}[ht]
\centering\includegraphics[width=150mm]{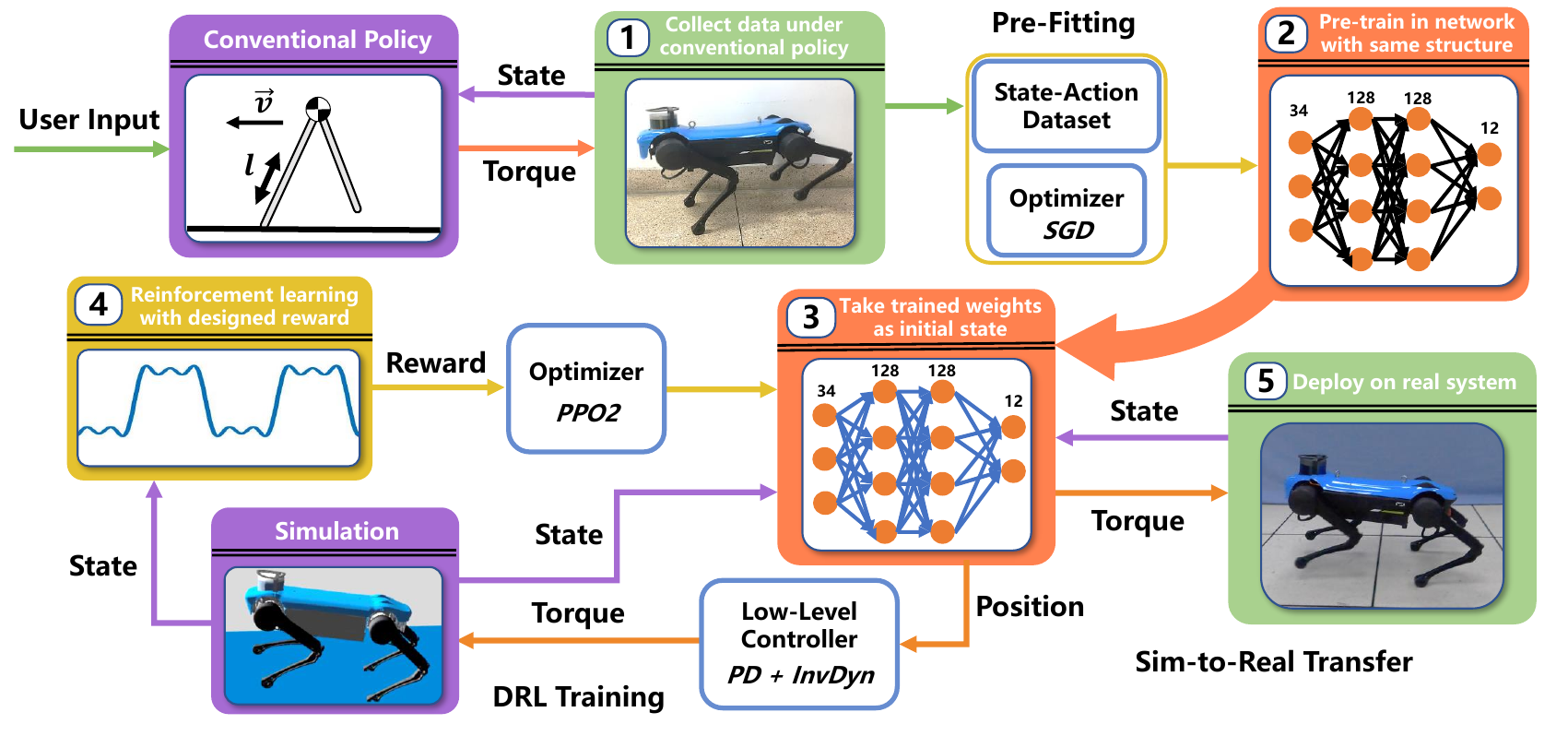}
\caption{Control architecture and workflow of training and deployment, including two parts: pre-fitting, and deep reinforcement learning. After training, the model is deployed on the real robot directly.}
\vspace{-4mm} 
\label{mainworkflow}
\end{figure}

\subsection{Main Workflow of DRL} The main workflow is shown in Fig. \ref{mainworkflow}. The scratch NN is trained on a dataset gathered from low-cost model-based policy with supervised learning. Then we initialize the policy in main DRL loop with pre-fitted NN's parameters. The conventional policy and pre-fitting algorithms are introduced in section \ref{pfDRL}.

In the main training loop in simulation, the state vector is retrieved from the simulation environment. It is a 34-dimensional vector comprising 1-dimensional body height, 3-dimensional body orientation, 3-dimensional body linear velocity expressed in body frame, 3-dimensional body angular velocity expressed in body frame, 12-dimensional joint position and 12-dimensional joint velocity. Clock signal is excluded, so the controller is not time-based.

The state vector is then sent to the deep neural network (DNN) as a frame of input, and the DNN instantly generates a corresponding frame of action consisting of a 12-dimensional output vector, which represents the expected joint positions.
A stable PD controller was used to convert the expected joint position into a 12-dimension expected joint torque vector. In addition to the basic PD torque, calculated with the joint position and velocity, an array of feedforward torques was added to the torque vectors to stabilize the robot system and help the joints move to the expected position more accurately. 

The feedforward torques are expected torques calculated with the expected states of the joints with floating-base inverse dynamics compensation where the acceleration is calculated with the difference of velocity. The torque vector is sent to the robot in the simulation environment as commands. The work loop then repeats itself. When training the neural networks, input and output data vectors along with the reward values are stored in tuples for calculating the neural networks' weight offsets.

After the simulation training procedure, the neural network, which can perform well, is ready to be applied on an actual robot. The same structure, except for the training part, is deployed on the actual robot for offline running. Because of the difference between simulation and real robot systems, the real robot will be able to bound after the bridging technique mention in section \ref{secBridge}.

\subsection{Reinforcement Learning}The motion state of the robot at a specific time is constrained by the previous state, therefore, the locomotion control is a Markov decision process, which is suitable for reinforcement learning (RL). The action  performed by the robot using policy influences the probability distribution of the state transition, and the result of the transition leads to a corresponding reward value , which implies how successful the state is. Hence, optimizing the performance of the controller is equivalent to maximizing the discounted reward function.

\begin{equation}
\pi^{*}(\theta)=\underset{\theta}{\operatorname{argmax}} \mathbb{E}_{\tau(\pi(\theta))}\left[\sum_{t=0}^{\infty} \gamma^{t} r_{t}\right]
\label{eq:expectReward}
\end{equation}
where $\gamma$ is the discount factor, which implies that the importance of the reward value drops as time passes. In addition, $\theta$ refers to the parameters in the policy; in our work,  $\theta$ represents the weights in the DNN. To train the DNN controller more efficiently, we chose Proximal Policy Optimization algorithm\cite{schulman2017trust, schulman2017proximal}, which is based on Actor-Critic structure\cite{article}, in our training process. 

\subsection{PF-DRL (Pre-Fit Deep Reinforcement Learning)}
\label{pfDRL}
There are two multi-layer perceptrons (MLP) deployed in the PPO algorithm, named the actor and critic. The structures of the neural networks are shown in Fig. \ref{mainworkflow}. With the random initialized weights, the most common hacking is in-place hind-leg trembling. Instead of wasting effort tuning reward coefficients, we deployed the pre-fitting method, which is efficient and often used in deep learning to solve this problem\cite{LDArch_forAI}, and meanwhile prevent introducing reference to final implementation on actual robots.

As for the pre-fitting data, we turned to model-based control for help. We recorded data corresponding to state space and action space, when the robot constrained under conventional model-based policies. The model-based policy used for data gathering is proposed by Raibert\cite{RaibertLRTB86}. It adopts a reduce-order model, and divides locomotion task into three sub-tasks which are velocity tracking, attitude control and orientation control. Velocity target tracking is realized by selecting foothold position according to neutral point theory and switches between phases according to a finite state machine.

After the data was collected, we constructed the neural network model with the same structure as the actor net of the RL model. We divided the recorded data as a training set and validation set. The input data of the neural network comprises 34 values recorded at the same time and the label of each input includes the 12 target positions by the conventional controller after 0.01 s, which is the control frequency in the RL simulation. We trained the neural network with different optimizers and learning rates in turn to minimize the mean squared error loss (Table \ref{Training Sequence}).  By using this hybrid training strategy, we can take advantage of high convergence speed of Adam and good generalization performance of SGD. Using single optimizer with a decaying learning rate can lead to similar result but extra tuning will be needed. 

\begin{figure}[ht]
  \centering
  \includegraphics[scale=0.54]{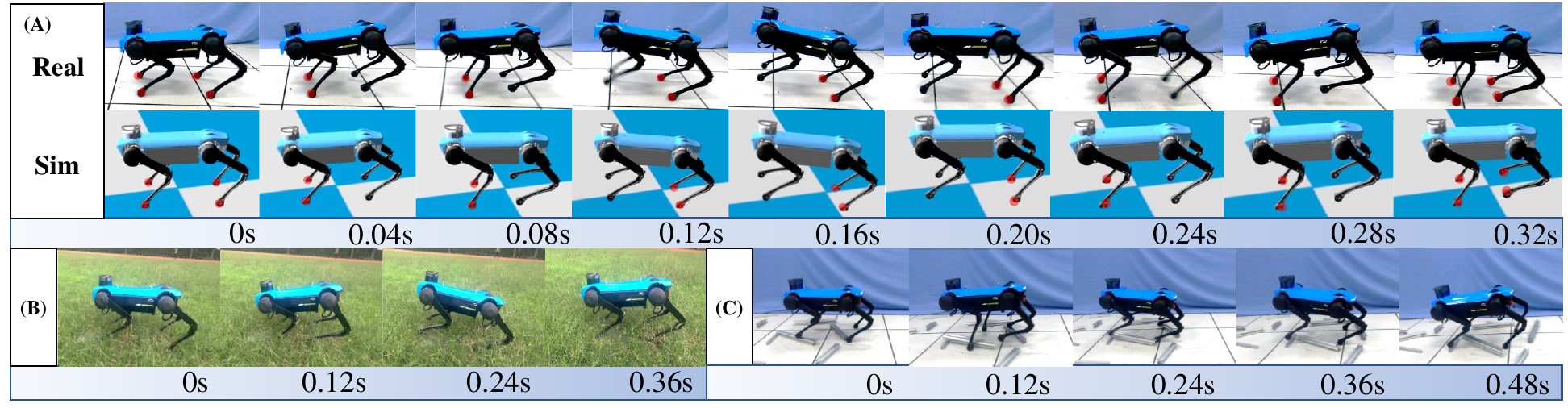}
  \caption{Quadruped bounding in simulation and in the indoor and outdoor environments. (A) The comparison between simulation and real world implementation, red dots highlight the contact points with the ground. (B) and (C) Traversal over rough grass field and plastic floor with sparse obstacles.}
  \vspace{-4mm} 
  \label{Snapshot}
\end{figure}

\begin{table}[tb]
    \caption{Training Sequence}
    \label{Training Sequence}
    \centering
    \begin{tabular}{c c c}
    \toprule[2pt]\text { optimizer } & \text { Learning rate } & \text { Training times } \\
    \hline \text { SGD } & 1e-2 & 500 \\
    \text { Adam } & 1e-3 & 500 \\
    \text { Adam } & 1e-4 & 500 \\
    \toprule[2pt]
    \end{tabular}
\end{table}

When the trained network converged, we transferred its weights into the actor net of the RL model as the initialization and started the training process directly.

\subsection{Design of Reward Function}
For RL processes, reward function is a vital factor in training. The RL algorithms will automatically find the trajectory that maximizes the total reward value. A proper reward function can improve the learning efficiency substantially. In our work the reward function comprises positive terms related to the main purpose of the movement such as the planar velocity of the whole robot, negative terms (also called cost terms) to regulate the locomotion, and negative terms to restrict the energy cost and improve the safety of the robot (Table \ref{Reward Terms}).

\begin{table}[!h]
    \caption{Reward Terms}
    \label{Reward Terms}
    \centering
    \setlength{\tabcolsep}{1mm}{
    \begin{tabular}{c c c}
    \toprule[2pt]\text { Reward } & \text { Formula } & \text { Coefficient Value } \\
    \hline \text { Body Velocity } & $k\left(\left|v_{x}^{I}\right|^{2}+\left|v_{y}^{I}\right|^{2}\right)$ & k=160.0 \\
    \text { Joint Torque } & $k \cdot \tanh (c t) \sum \tau_{i}$ & k=-0.002, c=0.04 \\
    \text { Joint Velocity } & $k \cdot \tanh (c t) \sum \dot{q}_{i}$ & k=-0.0003, c=0.02 \\
    \text { Gait } & $k \sum_{i=0}^{3} S(t+\delta_i) G(i, t)$ & k=-50.0 \\
    \text { Position Uniformity } & $\left(\left|q_{L F}-q_{R F}\right|+\left|q_{L H}-q_{R H}\right|\right)$ & k=-0.01 \\
    \text{ Torque Uniformity } & $k\left(\left|\tau_{L F}-\tau_{R F}\right|+\left|\tau_{L H}-\tau_{R H}\right|\right)$ & k=-0.001 \\
    \text { Smoothness } & $k\|\boldsymbol{\tau}(t)-\boldsymbol{\tau}(t-d t)\| $& k=-1 e-6 \\
    \text{ Pitch Limitation } & $k(|\phi|) \text { when }|\phi|>0.3$ & k=20.0 \\
    \toprule[2pt]
    \end{tabular}}
\end{table}
\noindent where $v_x^I$ refers to the speed of the robot base on axis $x$ under frame $I$, $q_i$ and $\tau_i$ refer to the position and torque of joint $i$, respectively, $\phi$ refers to the pitch angle of robot torso, $\tau_i$ is the 12-dimensional torque vector at time t,  $\omega$ is the frequency of desired gait, and $\delta_i$ is phase offsets of each legs. In the case of bounding, $\delta_forelegs = 0$, $\delta_hindlegs = \frac{\pi}{2\omega}$. $S(t)$ and $G(i, t)$ are special functions that can be described as formula \eqref{eq:Flourier3} and \eqref{eq:Gaitflag}.
\begin{align}
    \text{S(t)} &= \sin{\omega \text{t}} + \frac{1}{3} \sin{3\omega \text{t}} + \frac{1}{5} \sin{5\omega \text{t}} \label{eq:Flourier3}\\
    \text{G(i, t)} &= \left\{
    \begin{array}{l}
         +1 \quad \text{When foot i touches ground}  \\
         -1 \quad \text{When foot i does not touch ground}
    \end{array}
    \right.
    \label{eq:Gaitflag}
\end{align}

\begin{figure}[h]
  \centering
  \includegraphics[width=10cm]{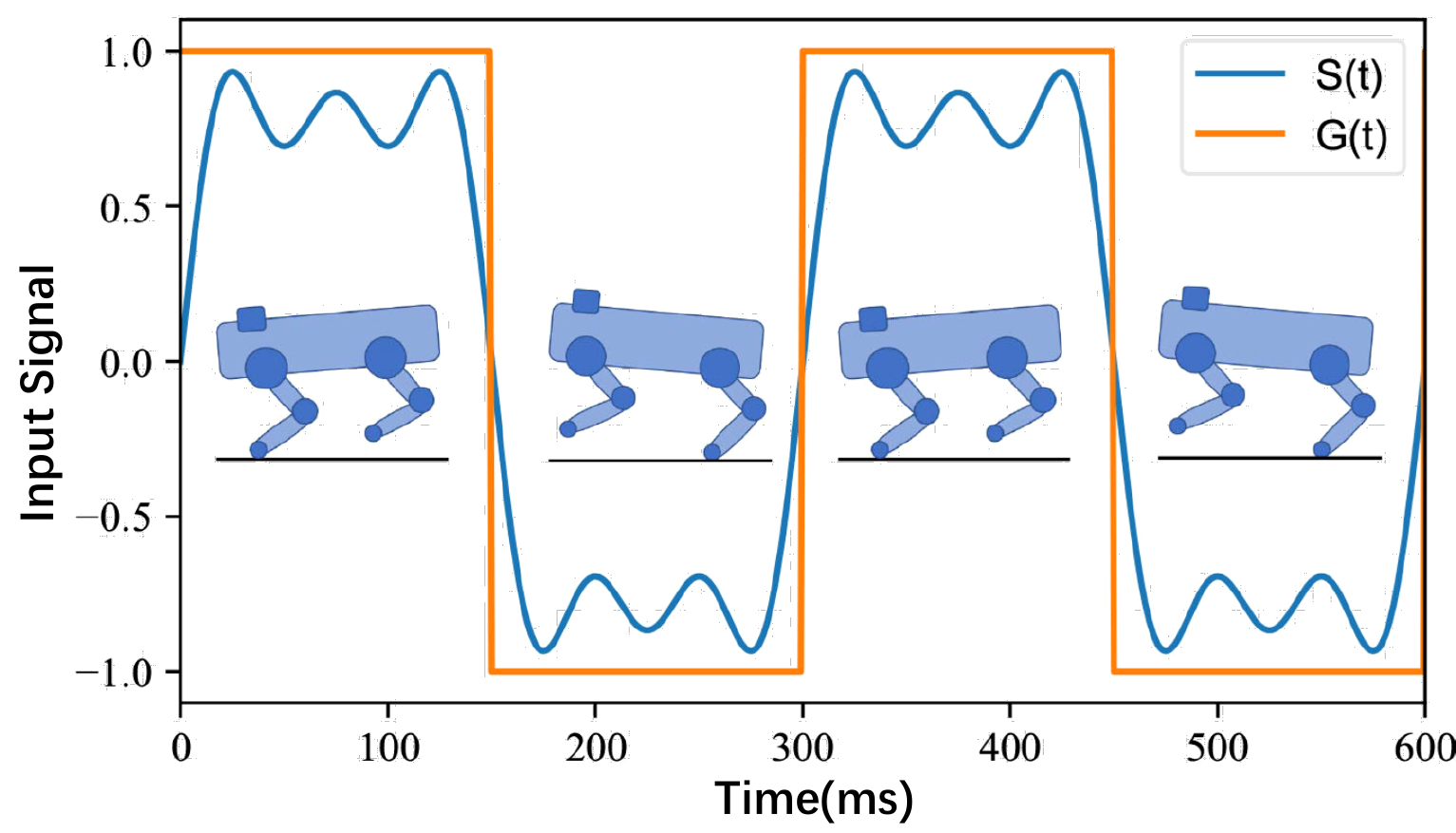}
  \caption{Illustration of contact states and reward. $G(i,t)$ represents current contact state of the robot and $S(t)$ indicates whether that foot should touch the ground. The plot shows example signals of forelimbs when bounding with a period of 0.3s, and an illustration of an ideal case of the robot in motion in order to achieve the maximum reward in the \textit{Gait Reward} term. }
  \label{Gait Signal Plot}
\end{figure}

Meanwhile, to train the robot to perform the desired gait, we designed a special gait signal to instruct the robot to take steps at a specific frequency. The gait signal is periodic and zero-symmetric, and it uses a signed term to convert the foot contact state to a reward value. If the foot touches the ground at a undesired time, the negative gait signal will lead to a negative reward and increase the cost (see Fig. \ref{Gait Signal Plot}). In our work, we use a third-order superposition of trigonometric functions. This form of signal has a larger root mean square value (RMS) and is still differentiable, which can avoid the risk of non-convergence when using simple rectangle signals.

\begin{figure*}[htb]
  \centering
  \includegraphics[scale=0.44]{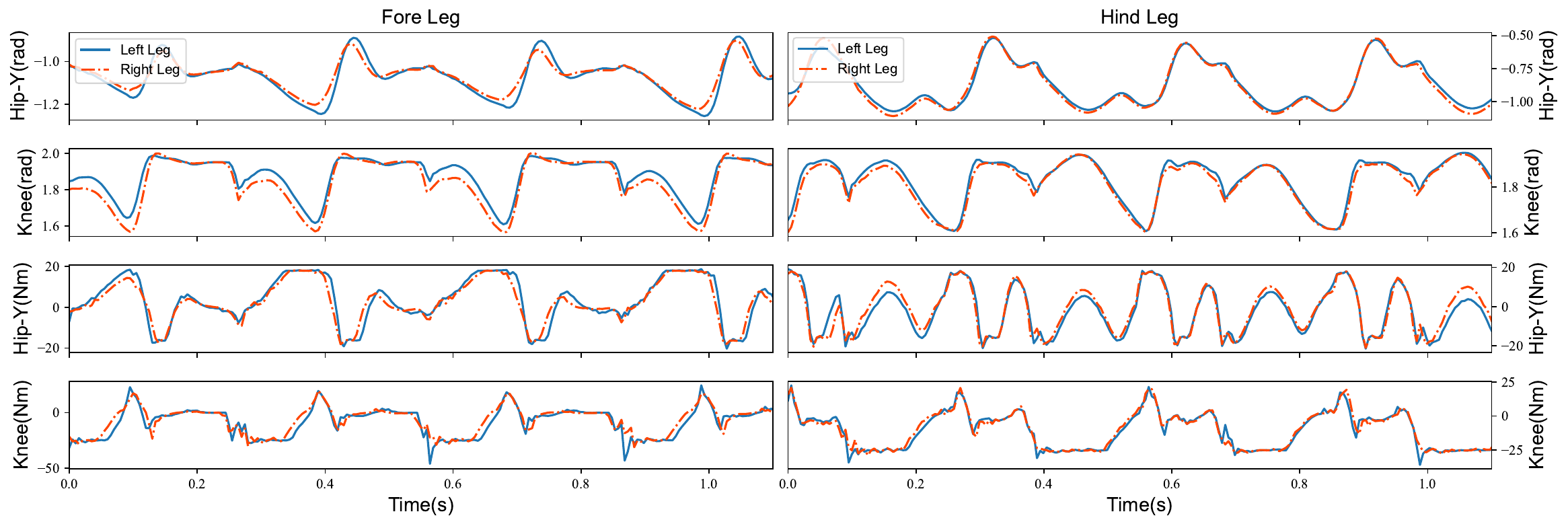}
  \caption{Experimental data of joint positions and torques during bounding, showing the symmetry between left and right legs.}
  \vspace{-5mm} 
  \label{Runtime Plot of Joint Positions and Torques}
\end{figure*}

\begin{table}[bt]
    \caption{Technical Specification of Jueying Mini}
    \label{Technical Specification of Jueying Mini}
    \centering
    \begin{tabular}{c c c}
    \toprule[2pt] \textbf { Property Name } & \textbf { Value } & \textbf { Unit } \\
    \hline \text { Body Size } & 0.7 * 0.4 * 0.5 & \multirow{2}{*} {m} \\
     \text { Leg Length(thigh+shank) } & 0.22+0.25 & \\
    \hline \text { Hip Roll Joint Position } & $-22.0 \sim 22.0$ & \multirow{3}{*} {deg}\\
     \text { Hip Pitch Joint Position } & $-158.0 \sim 28.0$ & ~\\
     \text { Knee Joint Position } & $38.0 \sim 163.0$ & ~\\
    \hline \text { Hip Roll Joint Velocity } & $-15.0 \sim 15.0$ &\multirow{3}{*}{rad/s} \\
     \text { Hip Pitch Joint Velocity } & $-18.0 \sim 18.0$ & ~ \\
     \text { Knee Joint Velocity } & $-20.0 \sim 20.0$ & ~\\
    \hline \text { Hip Roll Joint Torque } & 10.0 (\text {Peak} 23.0) & \multirow{3}{*}{Nm}\\
     \text { Hip Pitch Joint Torque } & 10.0 (\text {Peak} 26.0) & ~ \\
     \text { Knee Joint Torque } & 17.0 (\text {Peak} 41.5) & ~ \\
    \bottomrule[2pt] 
    \end{tabular}
\end{table}

\subsection{Bridging the Sim-to-Real Gap}
   \begin{table}[t]
    \caption{Noise Coefficients}
    \label{Noise Coefficients}
    \centering
    \begin{tabular}{c c c}
    \toprule[2pt]\text { Noise } & \text { Standard Deviation } & \text { Unit } \\
    \hline \text { Link Mass } & $\pm 5$ & \% \\
    \text { Link Inertia } & $\pm 10$ & \% \\
    \text { Link CoM } & $\pm 7.5$ & cm \\
    \text { Ground Friction } & $\pm 0.1$ & \\
    \text { Ground Restitution } & 0.15 & \\
    \toprule[2pt]
    \end{tabular}
    \end{table}
    
There are multiple gaps between the simulation environment and physical robots. We identify the main factors are the measurement errors when the systems is modeled. 

    1. To overcome uncertainty in the robot model, we added stochastic noise to both the environment and robot parameters. The terrain in the simulation was also randomized. The noise coefficients are shown in Table. \ref{Noise Coefficients}.

    2. Randomize the initial head direction every episode. The yaw angle is sampled from a uniform distribution $U(-\pi, \pi)$. This is to prevent over-fitting to the head direction and terrain conditions.
    
    3. To achieve high-frequency control in real time on the robot, we transform the weights of neural network in CSV format and compute forward network by C++.
    
    4. We lower the joint proportional gain to achieve better sim-to-real transfer, which is proved that lower proportional gain on real robot can make the joints behave like a torque controller and thus lead to agile performance.

\label{secBridge}

\section{Platform Overview}

\subsection{Robot Platform}

\begin{figure}[!h]
\centering
\includegraphics[width=10cm]{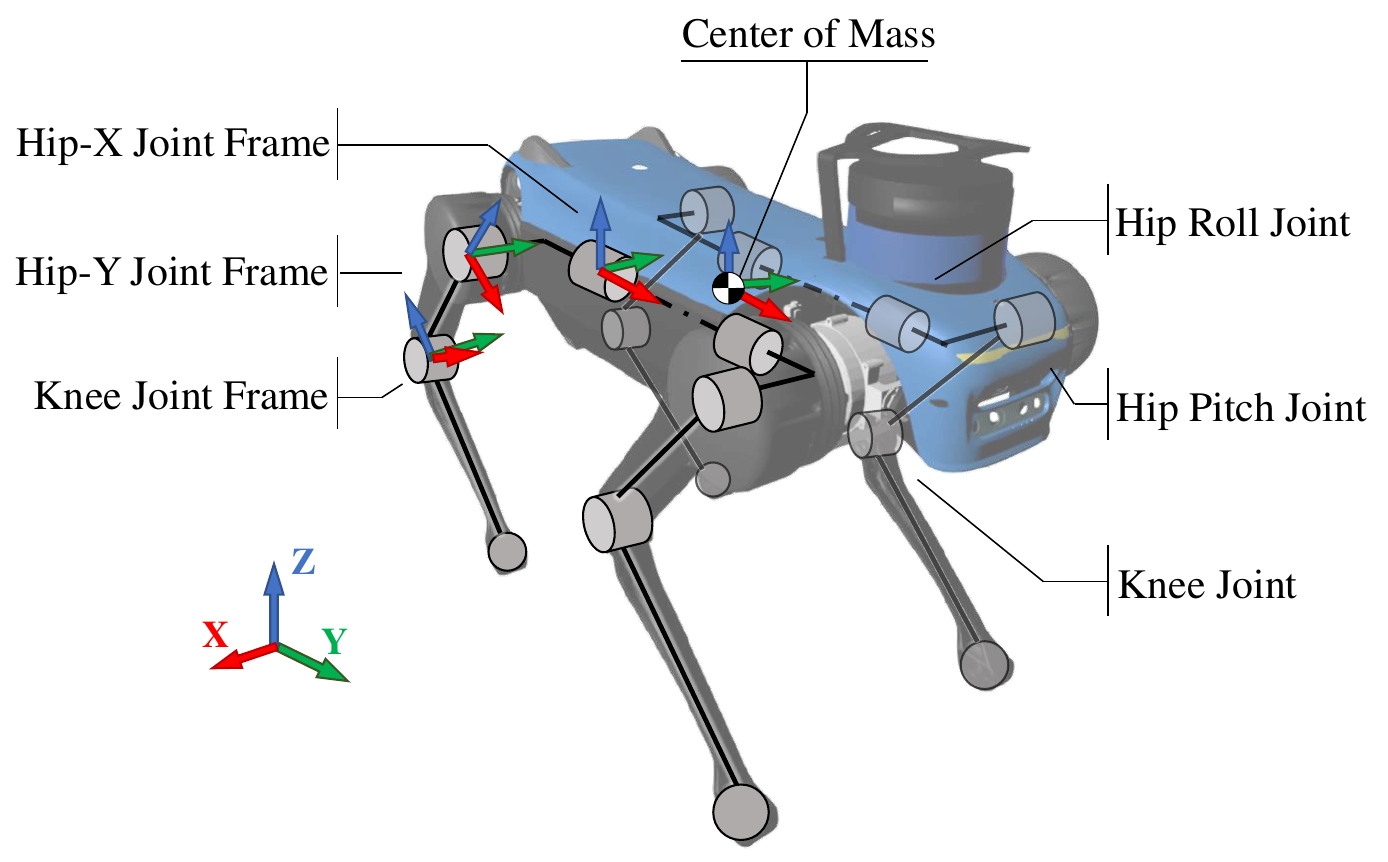}
\caption{Jueying Mini robot and its kinematic configuration.}
\label{platform}
\end{figure}

Our work is deployed on the quadrupedal robot Jueying Mini (see Fig.\ref {platform}). The Jueying Mini is a 12-DOF quadruped robot that focuses on agility. With a weight of 22 kg and joints actuated by brushless electric motors, the Jueying Mini can perform various movements. Its technical specifications are listed in Table \ref{Technical Specification of Jueying Mini}.

\subsection{Modeling}

To simulate the robot, we simplified it into a joint-link system comprising rigid links and revolute joints. The coordinates of each link are located on its parent joint. The directions of all coordinates are the same when the robot is in the initial position. A simplified model of the robot and its coordinates are shown in Fig. \ref{platform}.

\subsection{Simulation Environment}

We adopted RaiSim \cite{8255551} as the dynamic simulation environment. Because of its unique method of calculating contact forces, RaiSim is much faster than other dynamic simulation software. The simplified model of the Jueying Mini was deployed in the RaiSim environment with the corresponding controllers.

\section{Results }

\subsection{Training and Testing in Simulation}

\begin{figure}[h]
  \centering
  \includegraphics[width=80mm]{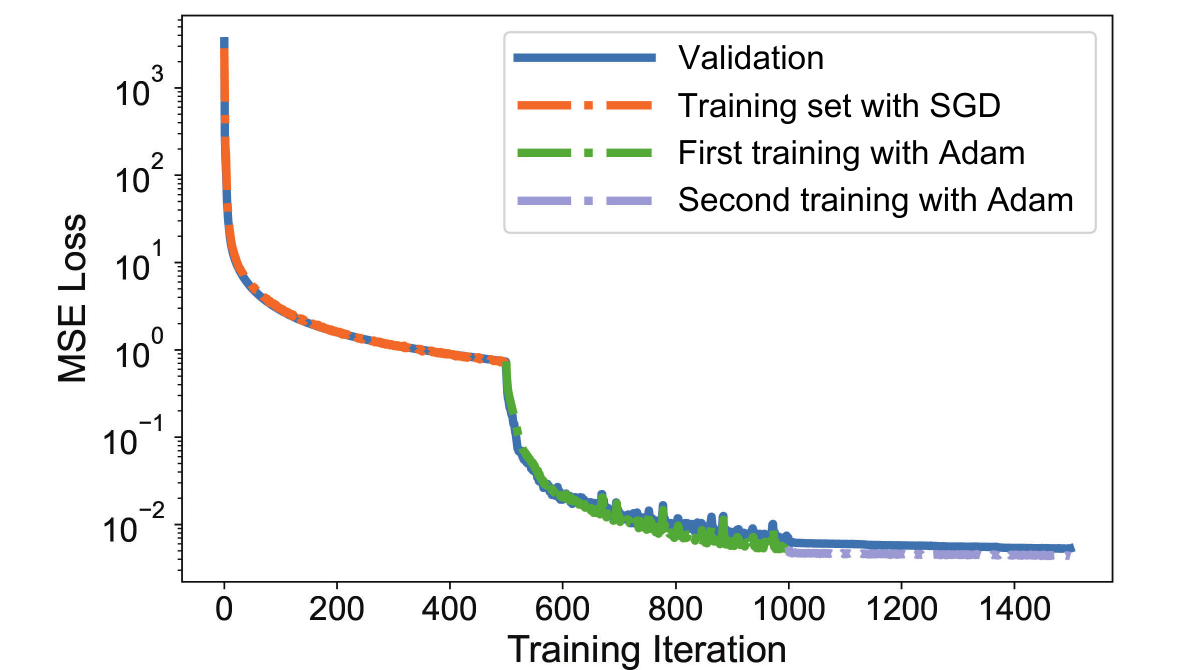}
  \caption{Training losses in the pre-fitting stage.}
  \vspace{-4mm} 
  \label{Training loss}
\end{figure}

We applied the deep pre-fitting methods on the actor network under PyTorch \cite{NEURIPS2019_9015}, and the training loss is shown in Fig. \ref{Training loss}. With introduced pre-fitting method the mean squared error loss was reduced  from $10^4$ to $4\times10^{-3}$ on the training set after 1,500 iterations, and the network began to perform a preliminary bounding motion but with frequent falling as a warm-start policy. On this basis, our next step is to continue training in a DRL fashion to refine the policy, which can then achieve continuous and cyclic bounding. 

\begin{figure}[h]
  \centering
  \includegraphics[width=80mm]{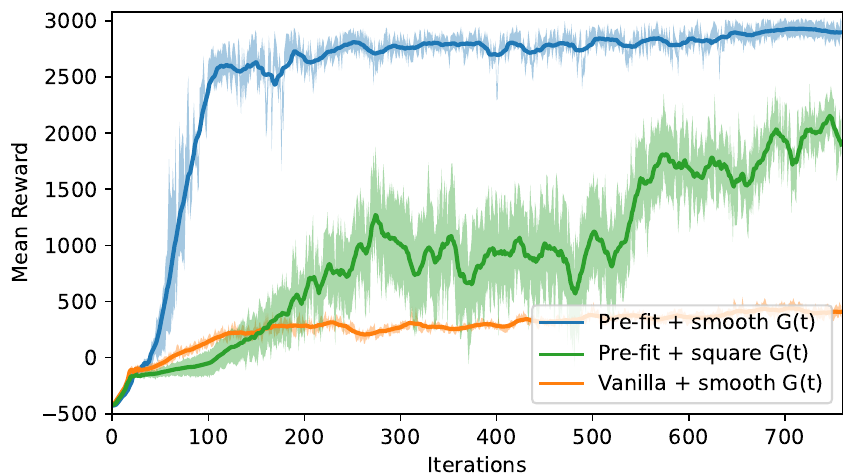}
  \caption{Rewards over episodes with and without pre-fitting.}
  \vspace{-4mm} 
  \label{Episode reward value}
\end{figure}

The pre-fitted actor network, or so-called policy network, was then put into simulation and optimized by DRL methods. We created 160 parallel RaiSim simulation environments that share the same policy network and train it synchronously speed up the data collection. The reward curves using different methods are shown in Fig. \ref{Episode reward value}. Instead of bounding, the policy without pre-fitting is stuck in place and thus receives low and stable rewards compared to the other methods. The pre-fitted policy using square gait signal performs bounding after training, but training takes longer and its reward is less stable while training. Combining both pre-fitting and smooth gait signal, the policy learns to bound in less iterations. 

\begin{figure*}[tb]
  \centering
  \includegraphics[scale=0.74]{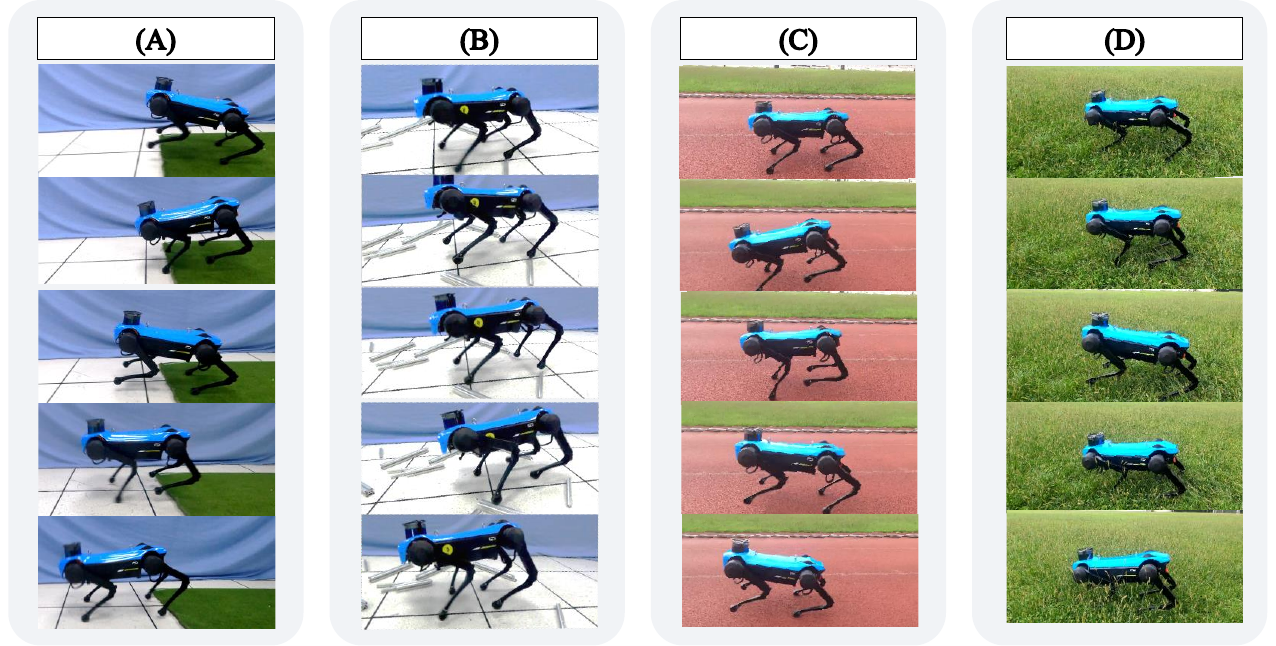}
  \caption{Jueying Mini robot bounding in various testing scenarios using the same PF-DRL trained policy. (A) Smooth transitions between regions with different properties (friction, stiffness). (B) Restoring balance after stepping on slippery objects twice. (C) Traversing soft outdoor ground. (D) Robust traversal on uneven grass field with shallow pits.}
  \vspace{-5mm} 
  \label{showcase}
\end{figure*}

\subsection{Real Robot Implementation }

After the two training steps above, the robot performed the desired gait on a flat terrain in simulation. The joint position, velocity, and torque are shown in Fig. \ref{Joint position, torque and gait on real robot}. Here, we take the left hind leg of the Jueying Mini as an example. In the chart, it is apparent that every data plot has a frequency of 3 Hz. The step frequency is consistent with the pre-fitting dataset.

\begin{figure}[h]
  \centering
  \includegraphics[scale=0.42]{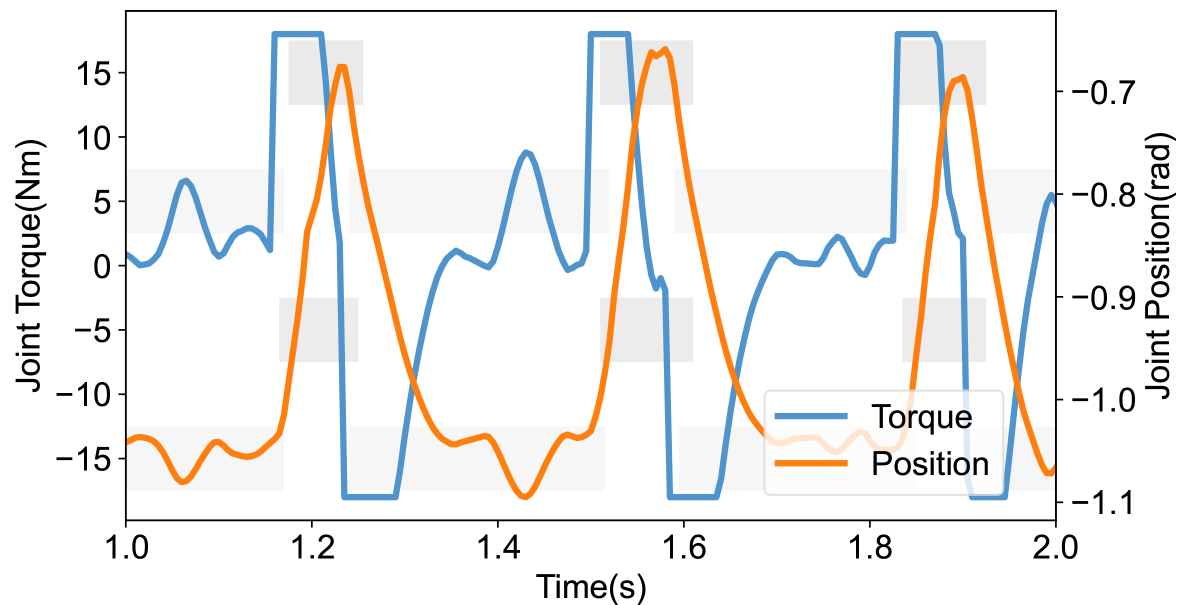}
  \caption{Smooth joint position and torque during bounding and gray shades are the estimated contact states.}
  \vspace{-3mm} 
  \label{Joint position, torque and gait on real robot}
\end{figure}

\begin{figure}[h]
  \centering
  \includegraphics[scale=0.51]{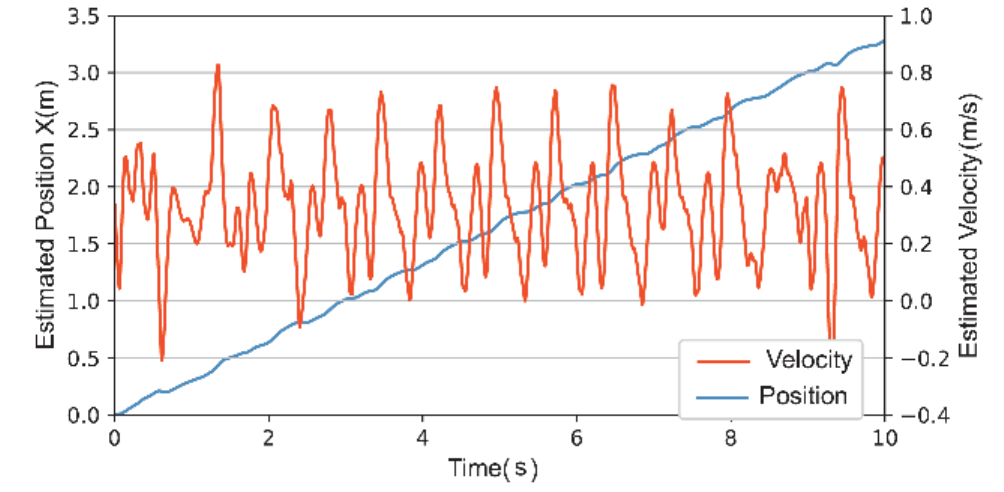}
  \caption{Position and forward velocity of the robot torso.}
  \vspace{-4mm} 
  \label{positionX&velocityX}
\end{figure}

When the policy network was optimized and able to perform a continuous and agile bounding gait in simulation, it was ready to be transferred to real robots. The policy MLP was deployed on a physical Jueying Mini robot and reached a peak speed of 0.75 m/s on flat indoor ground, with an average speed of 0.32m/s, shown in Fig. \ref{positionX&velocityX}. The gait performed in the real world was consistent with the simulated gait. The corresponding data is shown in Fig. \ref{Runtime Plot of Joint Positions and Torques}. Meanwhile, the controller showed sufficient robustness and ability to recover balance. The robot with the trained policy network was able to bound through unstructured terrain such as grass field and step over small obstacles about 4 cm in height (Fig. \ref{Snapshot}(b), Fig. \ref{showcase}(d)). See more details in \textit{the accompanying video}.

With the same policy, the physical robot can also bound on unseen complex discrete terrain. In the experiment, we put aluminum profiles of different shapes on the ground. The robot could bound through successfully (Fig. \ref{Snapshot}(c), Fig. \ref{showcase}(b)). In other words, when the contact point of the foot landed on an aluminum profile and negatively influenced the pose of the robot, the robot would adjust its pose to avoid slipping. This means that the policy we trained with RL is robust and has strong generalization ability. 

\begin{figure}[h]
  \centering
\includegraphics[scale=0.52]{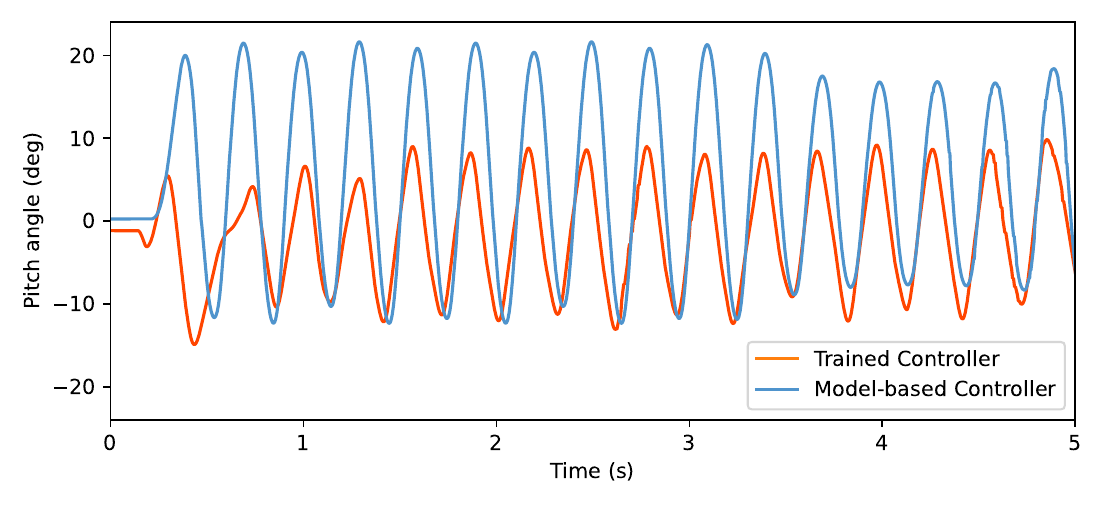}
  \caption{Pitch angle of the robot torso.}
  \vspace{-5mm} 
  \label{pitchPlot}
\end{figure}

On contrast, the model-based controller used for data generation failed to pass terrain with 4cm metal profile. When the robot tries to lift its rear legs while fore legs stepping over an obstacle, the robot crashes its head onto the ground and then loses balance. It makes sense because if the robot makes contact with parts other than feet, the unexpected contact force may severely violate the decision of model-base controller. We attribute this failure to the pitch angle fluctuation of model-based policy. Figure \ref{pitchPlot} plots pitch angles of both policies move forward on solid flat ground. Model-based policy has higher positive peak pitch value than trained policy, which means the robot's head would be much lower upon reaching the peak pitch. We come to the conclusion that trained policy improves traversability on complex terrains and reduces risk of unexpected contacts by performing a safer gait with less fluctuations.

\begin{figure}[h]
  \centering
\includegraphics[scale=0.48]{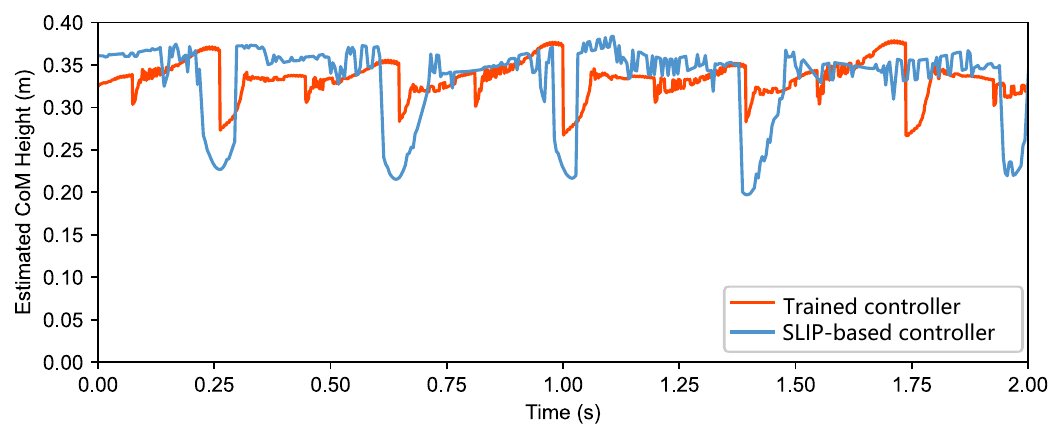}
  \caption{Height of CoM from the robot state estimation.}
  \vspace{-5mm} 
  \label{CoM height using different controllers}
\end{figure}

Despite of pitch angles, the trained neural network performs more smoothly when bounding in terms of body height. With trained policy, the height of the CoM changes in the range [0.25 m, 0.38 m] with a standard deviation of 0.02, whereas it fluctuates in [0.19 m, 0.41 m] with a standard deviation of 0.04 using the model-based controller used for pre-fit data generation. The comparison plot is shown in Fig. \ref{CoM height using different controllers}.

\section{Conclusions}

This paper presented a control method based on DRL for the Jueying Mini robot which can achieve a bounding gait both in simulation and on the real physical robot by a direct deployment of a learned neural network policy. The proposed method can train a robust control policy that can bound blindly on different indoor and outdoor terrains.

With the proposed pre-fitting method, we can transfer the conventional controller to a NN first, as a means of warm-start, and then further improve the control policy via RL training. By changing the coefficients of each reward, we were able to adjust the gait frequency more easily than using a conventional control method. 

Future work will consider extensions of aiding PF-DRL with environmental perception. In this study, we did not use camera or LiDAR systems, but they could provide more precise localization of the robot and provide navigation for bounding and traversal in different environments.  

\section{Funding}This work was supported by the National Key R\&D Program of China (Grant No. 2020YFB1313300) and Key Research Project of Zhejiang Lab (Grant No. 2021NB0AL03).

\section{Conflicts of Interest}The authors declare no conflict of interest.

\section{Acknowledgment}First, we wish to give our thanks to Chao Li, Xueyin Zhang, Feng Li, Xiaobo Mo, Zhen Chu and Chengxiao Li from DeepRobotics. As the logistics staff for Jueying Mini, they kept the hardware in good condition with their effort even after serious damage, and thus created the possibility for experiments on real robots. Second, we would like to express our gratitude to Yue Wu and Yan Xia, that helped with debugging and tuning learning algorithms. Their experience on software framework and learning systems prevented us from plenty of potential time-consuming design breaches.

\section{Bibliography}
\bibliographystyle{vancouver}
\bibliography{WileyNJD-VANCOUVER}

\end{document}